\documentclass{article}
\usepackage{fancyhdr}
\usepackage{neurips_2024}
\usepackage{graphicx} 
\usepackage{amsmath, amsfonts, amssymb}
\usepackage{booktabs}
\usepackage[utf8]{inputenc}
\usepackage{tikz}
\usetikzlibrary{shapes.geometric, arrows.meta, positioning, calc, backgrounds, fit}
\usepackage{xcolor}
\usepackage{wrapfig}
\usepackage[htt]{hyphenat}

\definecolor{processblue}{RGB}{0, 114, 189}
\definecolor{processgreen}{RGB}{119, 172, 48}
\definecolor{processorange}{RGB}{217, 83, 25}
\definecolor{processyellow}{RGB}{237, 177, 32}

\title{Motif-2-12.7B-Reasoning: A Practitioner's Guide to RL Training Recipes}
\author{\textbf{Motif Technologies}}

\begin{document}
\maketitle

\begin{abstract}
We introduce Motif-2-12.7B-Reasoning, a 12.7B parameter language model designed to bridge the gap between open-weight systems and proprietary frontier models in complex reasoning and long-context understanding. Addressing the common challenges of model collapse and training instability in reasoning adaptation, we propose a comprehensive, reproducible training recipe spanning system, data, and algorithmic optimizations. Our approach combines memory-efficient infrastructure for 64K-token contexts using hybrid parallelism and kernel-level optimizations with a two-stage Supervised Fine-Tuning (SFT) curriculum that mitigates distribution mismatch through verified, aligned synthetic data. Furthermore, we detail a robust Reinforcement Learning Fine-Tuning (RLFT) pipeline that stabilizes training via difficulty-aware data filtering and mixed-policy trajectory reuse. Empirical results demonstrate that Motif-2-12.7B-Reasoning achieves performance comparable to models with significantly larger parameter counts across mathematics, coding, and agentic benchmarks, offering the community a competitive open model and a practical blueprint for scaling reasoning capabilities under realistic compute constraints.
\end{abstract}

\begin{figure}[!h]
         \centering
         \includegraphics[width=1.0\linewidth]{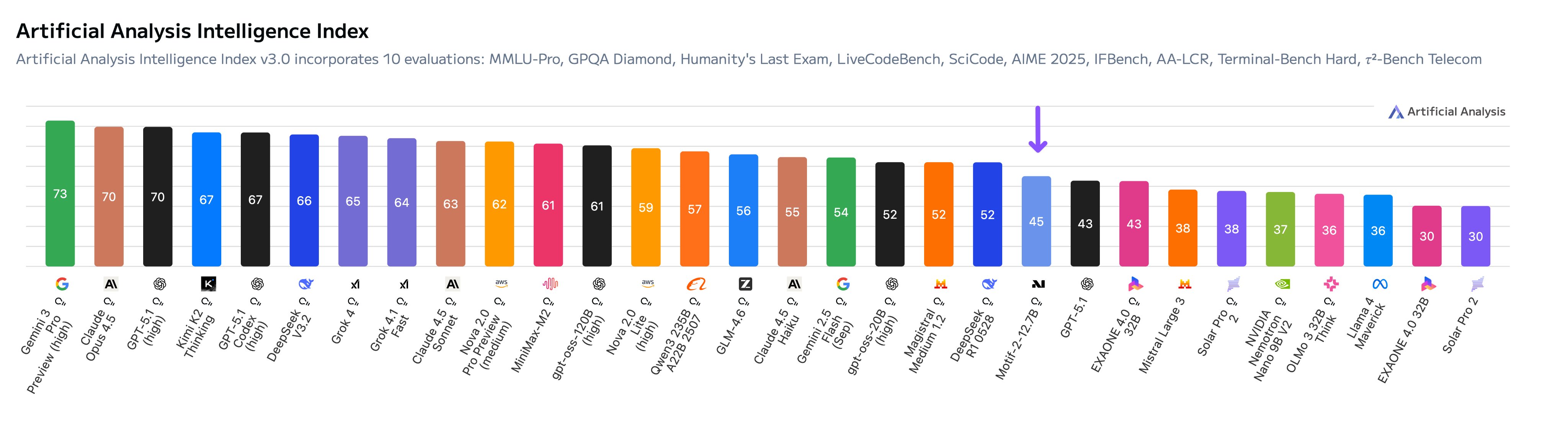}
         \caption{Motif-2-12.7B-Reasoning performance compared to the best proprietary and open-source models. The Artificial Analysis Intelligence Index (AAII) is a composite metric derived from the average scores of 10 diverse benchmarks. Despite its compact size of 12.7B parameters, motif-2-reasoning achieves performance comparable to larger-scale frontier models.}
         \label{fig:placeholder}
\end{figure}

\section{Introduction}
Recent advances in Large Language Models (LLMs) have delivered impressive gains in general instruction following, yet robust complex reasoning and long context understanding remain open challenges. A growing body of work shows that test-time scaling, which allocates more inference compute through longer thinking traces, multi-step deliberation, or multi sample reasoning, can substantially improve performance not only on mathematics and coding but also on instruction following, tool use, and agentic decision making. At the same time, both closed source frontier models and the strongest open weight systems increasingly take the form of reasoning optimized LLMs, as reflected in benchmarks such as the Artificial Analysis Intelligence Index (AAII)\footnote{\url{https://artificialanalysis.ai/}}, where “thinking” variants consistently outperform their non reasoning counterparts.

Despite this shift, training reasoning LLMs remains accessible to only a small number of organizations. Practical recipes for scaling reinforcement learning fine-tuning (RLFT), stabilizing training with long horizons, and efficiently handling long-context workloads are rarely documented in detail. In particular, while RLFT is often advertised as the next step beyond supervised fine-tuning (SFT), practitioners frequently observe that naïvely applying RLFT leads to model collapse or degradation on precisely the reasoning benchmarks we seek to improve. As a result, the community has strong evidence that reasoning models and inference scaling matter, but limited guidance on how to reliably train such models under realistic budget constraints.

In this context, we introduce Motif-2-12.7B-Reasoning, a 12.7B-parameter reasoning LLM trained on top of Motif-2-12.7B-Instruct~\cite{lim2025motif2127btechnical} via an additional stage of distribution-aligned SFT followed by a carefully engineered RLFT pipeline. Despite its relatively modest size, Motif-2-12.7B-Reasoning achieves performance comparable to, and in some cases surpassing, frontier models in the 30–40B parameter range across a broad suite of reasoning-heavy benchmarks. On AAII and related composite evaluations, Motif-2-12.7B-Reasoning attains higher scores than GPT-5.1, and among models that rank above it, none are smaller than 12.7B parameters.

We believe that genuine progress in AI relies on shared knowledge and reproducibility, not opaque “black-box” training pipelines. Our goal in this work is therefore twofold: to present Motif-2-12.7B-Reasoning as a competitive open-weight reasoning model, and to make its training process transparent and practically reproducible. We document not only the final recipe but also the design iterations, failure modes, and empirical lessons that shaped it, with an emphasis on decisions that materially affected stability, efficiency, and performance.

On the SFT side, we describe a two-stage, distribution-aligned curriculum that densifies and structures reasoning supervision. This includes targeted curation of multi-step reasoning data (math, code, and multi-hop natural language), alignment of SFT data distribution with the downstream RL objective, and curriculum strategies that gradually increase task difficulty and context length while preserving instruction-following fidelity.

On the RLFT side, we propose a practical RL recipe for reasoning and long-context use cases, focusing on techniques that can be implemented within realistic compute budgets. We detail our approach to reward modeling for diverse reasoning tasks, stabilization heuristics to avoid over-optimization and mode collapse, and scheduling strategies that balance exploration with preservation of base-model capabilities. We further analyze how these design choices interact with test-time scaling, showing that RLFT can be made both stable and reliably beneficial when applied with the right constraints.

Finally, we present comprehensive empirical results demonstrating that Motif-2-12.7B-Reasoning delivers strong gains over Motif-2-12.7B-Instruct and competitive baselines in mathematics, coding, instruction following, tool-augmented reasoning, and long-context benchmarks. We highlight its test-time scaling behaviour, its parameter-efficiency relative to larger frontier models, and its robustness across diverse evaluation settings. We hope that this report, by opening up both the model and its training methodology, empowers the open-source community to build upon these insights and to drive the next wave of advances in reasoning-centric LLMs.

\section{Part 1 -- System Optimization}
\label{sec:sys_ops}
\paragraph{Long Context Reasoning SFT}
Our SFT recipe requires extending the context length to 64K tokens, which in turn necessitates additional forms of parallelism beyond those used in pretraining. To address this, we adopt a hybrid parallelism strategy: within each node (intra-node) we adopt DeepSpeed-Ulysses~\cite{jacobs2023deepspeedulyssesoptimizationsenabling} sequence parallelism (SP) together with data parallelism with parameter sharding (DP-shard), while across nodes (inter-node) we use data parallelism with parameter replication (DP-replicate). In details, attention layers are handled with tensor parallelism (TP) and feed-forward networks (FFNs) operate under SP. Since FFNs using SP can be weight-sharded across the SP mesh, we construct a merged mesh that combines the intra-node DP-shard mesh with the SP mesh when sharding FFN parameters. Because Parallel Muon introduced at Motif-2-12.7B-Instruct~\cite{lim2025motif2127btechnical}  is designed to operate under arbitrary gradient and parameter placement configurations, no additional modifications to the optimizer were required for hybrid parallelization.

In addition, we applied fine-grained activation checkpointing rather than uniformly recomputing every decoder block. Instead of checkpointing at the block granularity, we analyzed the per-layer activation footprint and recomputation cost within each decoder block and manually tuned a more selective recomputation policy. This hand-optimized policy minimizes peak memory usage while avoiding unnecessary recomputation overhead.

These approaches enables efficient parameter distribution and alleviates memory pressure for long-context training. Using this configuration, we successfully trained models with 64K context length on H100 GPUs.\footnote{Our infrastructure is operated using the \texttt{SkyPilot}~\cite{yang2023skypilot} framework, which enables seamless multi-node and multi-cluster provisioning as well as unified resource abstraction across heterogeneous compute environments.}

\paragraph{Liger Kernel's loss function}
Reinforcement Learning (RL) requires substantially more memory than SFT, primarily because additional components such as the policy model must be kept in memory. Although the policy implementation we use\textemdash based on vLLM\textemdash supports features like sleep mode, these mechanisms do not fully eliminate its memory footprint. To address the additional memory pressure introduced during RL training, we adopt the Liger Kernel~\cite{hsu2025ligerkernelefficienttriton}'s loss function.

An examination of memory usage patterns in RL reveals that the peak consumption occurs immediately after the forward pass of the loss function and at the start of the backward pass. At this point, all checkpointed activations generated during the forward computation are still resident in memory; they are released only as the backward pass progresses. Notably, the logit activations, shaped as (\textit{context length}, \textit{vocabulary size}), are significantly larger than the typical activations of shape (\textit{context length}, \textit{hidden size}), meaning that the LM head and loss computation contribute a non-negligible portion of the memory footprint.

By using the Liger Kernel's loss function\textemdash which divides the activations along the context-length dimension, computes LM head liner on each shard, and computes loss and gradients of each shard\textemdash we can substantially reduce the memory pressure arising from these large logit tensors.
\section{Part 2 -- Reasoning SFT}
We introduce a reasoning-focused supervised fine-tuning (SFT) stage designed to strengthen multi-step reasoning, reinforce long-context consistency, and align the model with complex instruction-following capabilities.
Early experiments reveal that reasoning ability is highly sensitive to the composition of the dataset, reasoning depth, and mismatched reasoning patterns. These observations motivate a structured SFT process built upon twofold: (1) curriculum-based long-context adaptation and (2) distribution-aligned synthetic reasoning generation.

\subsection{Lesson from failures}
\paragraph{Lesson 1: Dynamic Dataset Distribution across Phases.}
Our empirical findings reveal that a static and uniform data distribution throughout the SFT process is suboptimal. Such a strategy often forces premature convergence, preventing the model from scaling its reasoning capabilities and risking catastrophic forgetting. Instead, a dynamic shift in the distribution of the dataset is essential. We structure this curriculum into two strategic stages:

\begin{itemize}
    \item \textbf{Stage 1 (Reasoning Foundation):} It focuses on establishing comprehensive competence across code, mathematics, STEM, and tool-use domains. We construct our training corpus by incorporating various open sources, including \texttt{Nemotron-Post-Training-Dataset}~\cite{NemotronPostTrainingDatasetV1, NemotronPostTrainingDatasetV2}, \texttt{OpenReasoningDataset}~\cite{moshkov2025aimo2, ahmad2025opencodereasoning, ahmad2025opencodereasoningiisimpletesttime}, and \texttt{Mixture-of-Thoughts}~\cite{openr1, penedo2025codeforces, lozhkov2025openr1math220k}. To ensure data quality, we apply the post-processing pipeline that filters samples based on difficulty, sequence length, and verification. For instance, in processing the \texttt{rstar-coder}~\cite{liu2025rstarcoderscalingcompetitivecode} dataset, Driven by empirical observations, we adhere to \textit{quality-over-quantity} principle. We isolate problems exhibiting discriminative difficulty by selecting cases where the pass count was strictly within the intermediate range ($0 < n_{\text{pass}} < 16$), thereby pruning both trivial and empirically unsolvable samples. From this subset, we retained only the trajectories that were explicitly verified and successfully executed (\texttt{verified=True} $\land$ \texttt{is\_passed=True}), prioritizing high-fidelity reasoning over data volume. This stage acts as a stabilizer, grounding the model's general reasoning capabilities while initiating exposure to medium-length contexts (16K--32K tokens).

    \item \textbf{Stage 2 (Deep Reasoning Specialization):} It targets complex inferential gaps by injecting high-granularity synthetic data, including Chain-of-Thought (CoT) intensive and failure-driven correction sets. Notably, we regenerate the reasoning traces for these samples to enforce structural alignment with the target model's reasoning distribution. Furthermore, this stage completes the context curriculum by extending the window to 64K tokens, enabling the model to sustain coherent reasoning over long sequences.
\end{itemize}

\begin{wrapfigure}{r}{0.35\textwidth} 
    \centering
    \begin{tabular}{lc}
        \toprule
        \textbf{Configuration} & \textbf{Pass@1} \\
        \midrule
        Baseline & 51.78 \\
        + \texttt{seed-oss} & 63.69 ($\uparrow \textcolor{red}{11.91}$) \\
        + \texttt{gpt-oss} & 33.92 ($\downarrow \textcolor{blue}{17.86}$) \\
        \bottomrule
    \end{tabular}
    \caption{\textbf{Impact of reasoning distribution alignment evaluated on LiveCodeBench v5.} \texttt{seed-oss} and \texttt{gpt-oss} denote synthetic data generated by \texttt{seed-oss-36b} and \texttt{gpt-oss-120b}, respectively.}
\label{tab:mismatch_impact}
\end{wrapfigure}

\paragraph{Lesson 2: The Distribution Mismatch Problem.} A key insight from our early experiments is that the quality of synthetic data encompasses more than just the correctness; the alignment of the reasoning distribution is crucial. To quantify this, we evaluate the impact of reasoning alignment on LiveCodeBench v5~\cite{jain2024livecodebench}, as shown in Table~\ref{tab:mismatch_impact}. We compare the baseline against configurations generated with synthetic samples from \texttt{seed-oss-36b}~\cite{seed2025seed-oss} and \texttt{gpt-oss-120b}~\cite{openai2025gptoss120bgptoss20bmodel}. While \texttt{seed-oss} yields significant gains, \texttt{gpt-oss} caused distinct performance degradation. This contrast implies that performance depends on the compatibility of reasoning traces, not merely the volume of data.

We hypothesize that the degradation caused by \texttt{gpt-oss} stems from a complexity mismatch. The teacher model's reasoning traces likely exhibit granularity and structural complexity that diverge from the student model's intrinsic reasoning style. This discrepancy creates a \textit{distribution mismatch}, where the imposed reasoning patterns conflict with the model's learning process rather than reinforcing it. This suggests that synthetic supervision is most effective when derived from sources that are compatible with the target model's capacity.

\subsection{Recipe}
\begin{figure}[!h]
\centering
\resizebox{\linewidth}{!}{%
\begin{tikzpicture}[
    node distance=1.5cm,
    font=\sffamily\footnotesize,
    process_box/.style={rectangle, draw=processblue!80, fill=white, thick, align=center, minimum height=1cm, minimum width=2.5cm, rounded corners},
    filter_box/.style={rectangle, draw=processorange!80, fill=white, thick, align=center, minimum height=2cm, minimum width=3.5cm, rounded corners},
    stage_box/.style={rectangle, draw=processgreen!80, fill=white, thick, align=center, minimum height=1cm, minimum width=3.5cm},
    arrow/.style={-Latex, thick, color=darkgray},
    label_text/.style={font=\sffamily\bfseries\small, color=darkgray}
]

\node[label_text] (pipeline_title) {Strategy 1: Data Generation \& Verification Pipeline (Quality)};

\node[process_box, below=0.5cm of pipeline_title] (gen1) {\textbf{1. Query Generation}\\(Diverse Domains)};
\node[process_box, below=0.8cm of gen1] (gen2) {\textbf{2. Response Generation}\\(w/ Reasoning Traces)};

\node[filter_box, below=1cm of gen2] (verify) {
    \textbf{3. Verification Gate}\\
    \begin{tabular}{@{}l@{}}
        $\checkmark$ Consistency \& Factuality \\
        $\checkmark$ Code Execution \\
        $\checkmark$ Math Correctness \\
        $\checkmark$ Structural Validity
    \end{tabular}
};

\node[process_box, fill=white, below=1cm of verify] (final_data) {\textbf{Final Aligned Data Pool}};

\draw[arrow] (gen1) -- (gen2);
\draw[arrow] (gen2) -- node[right, font=\scriptsize] {Raw Data} (verify);
\draw[arrow] (verify) -- node[right, font=\scriptsize] {Filtered High-Quality Data} (final_data);

\node[label_text, right=3cm of pipeline_title, yshift=-2.cm] (curriculum_title) {Strategy 2: Curriculum Learning (Stability)};

\node[stage_box, below=0.5cm of curriculum_title, fill=processgreen!10] (stage1) {\textbf{Stage 1: 16K Context}\\(Base Reasoning)};
\node[stage_box, below=0.5cm of stage1, fill=processgreen!30] (stage2) {\textbf{Stage 2: 32K Context}\\(Extended Context)};
\node[stage_box, below=0.5cm of stage2, fill=processgreen!50] (stage3) {\textbf{Stage 3: 64K Context}\\(Long-range Consistency)};

\draw[arrow] (stage1) -- (stage2);
\draw[arrow] (stage2) -- (stage3);

\begin{pgfonlayer}{background}
    \node[fit=(pipeline_title)(gen1)(final_data), draw=gray!30, fill=gray!5, rounded corners, dashed, inner sep=0.3cm] (strat2_bg) {};
    \node[fit=(curriculum_title)(stage1)(stage3), draw=gray!30, fill=gray!5, rounded corners, dashed, inner sep=0.3cm] (strat1_bg) {};
\end{pgfonlayer}

\draw[arrow, line width=1.0mm, color=gray!45] 
    (strat2_bg.east |- strat1_bg) -- (strat1_bg.west) 
    node[midway, above, font=\bfseries\scriptsize, color=black, yshift=2pt] {Feeds into};
    
\end{tikzpicture}
}%
\caption{\textbf{Overview of the SFT Recipe.} The recipe combines a synthetic data generation and verification pipeline (left) to ensure reasoning quality and alignment, feeding into a progressive curriculum learning schedule (right) to ensure training stability across extending context lengths.}
\label{fig:sft_recipe_diagram}
\end{figure}
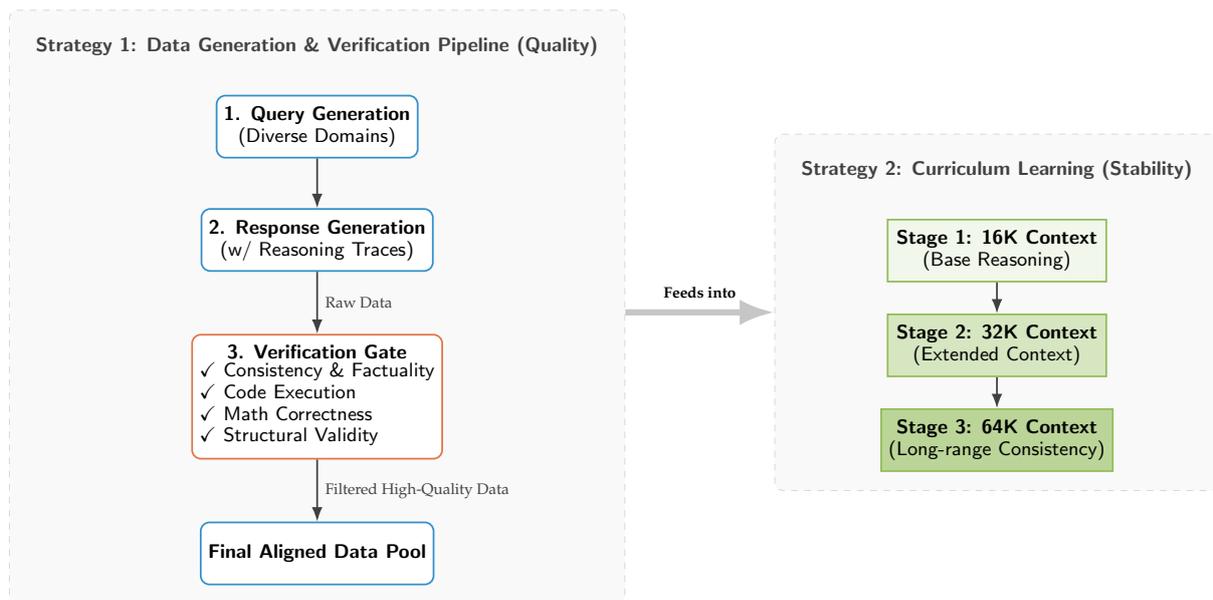

To address the challenges, we establish a robust SFT recipe focused on stability through curriculum learning and quality through distribution alignment, as illustrated in Figure \ref{fig:sft_recipe_diagram}.

\paragraph{Strategy 1: Synthetic Data Generation (Injecting Reasoning Signals)}
We generate diverse synthetic datasets specifically engineered to inject strong reasoning signals into target domains such as code generation, tool-calling, mathematics, and instruction following. Unlike standard datasets, these datasets emphasize structured, multi-step reasoning with explicit thought traces, enabling the model to internalize the logical progression required for complex problem-solving.

\paragraph{Structured Generation and Verification Pipeline}
To ensure data consistency and avoid the distribution mismatch problem, we construct a multi-stage pipeline: \textit{Query Generation $\to$ Response Generation $\to$ Verification}. The verification stage is critical for maintaining high data fidelity and employs a suite of automated checks:
\begin{itemize}
    \item \textbf{Consistency \& Factuality:} Verifying semantic alignment with the original query and validating factual accuracy.
    \item \textbf{Code Execution:} Running execution-based tests for programming tasks to ensure functional correctness.
    \item \textbf{Mathematical Correctness:} Verifying the accuracy of final answers.
    \item \textbf{Structural Validity:} Ensuring the reasoning trace maintains a logical and coherent structure aligned with our target distribution.
\end{itemize}

\paragraph{Strategy 2: Curriculum Learning (Progressive Context Extension)}
Instead of exposing the model to extremely long sequences abruptly, we adopt a progressive length-extension curriculum. The context window is expanded in stages: 16K $\to$ 32K $\to$ 64K tokens.
This gradual adaptation allows the model to learn long-range dependencies more effectively while mitigating the instability often associated with sudden context scaling. Furthermore, it optimizes attention behavior, enhancing the model's ability to maintain coherent reasoning across extended contexts.
\section{Part 3 -- Reasoning RL}

\subsection{Background}
Group Relative Policy Optimization~(GRPO)~\cite{shao2024deepseekmathpushinglimitsmathematical} is a critic-free variant of PPO tailored for reasoning tasks. Unlike standard actor-critic methods that rely on a separate value function, GRPO estimates the baseline directly from the group statistics of sampled outputs. Specifically, for each input prompt $q$, the model generates a group of outputs $\{o_i\}_{i=1}^G$, and the rewards are standardized within this group to compute the relative advantage. The GRPO objective is formally defined as:
\begin{equation}
\mathcal{L}_{\text{GRPO}}(\theta)
= \mathbb{E}_{q \sim P(Q), \{o_i\} \sim \pi_{\theta_{\text{old}}}}\left[
  \frac{1}{G} \sum_{i=1}^G \min\big(
    \rho_i(\theta) A_i,\,
    \operatorname{clip}(\rho_i(\theta), 1-\epsilon, 1+\epsilon) A_i
  \big)
\right],
\label{eq:grpo}
\end{equation}
where $\rho_i(\theta) = \frac{\pi_\theta(o_i\vert q)}{\pi_{\theta_{\text{old}}}(o_i\vert q)}$ denotes the importance ratio, and $A_i$ represents the advantage normalized using the group's mean and standard deviation. This formulation effectively encourages responses that outperform the group average without incurring the computational overhead of training a value network. Note that we employ GSPO~\cite{zheng2025groupsequencepolicyoptimization}, which utilizes a sequence-level formulation for the importance ratio $\rho(\theta)$.

\subsection{Lesson from failures}
Before initiating full-scale RL training, we first conduct a series of small-scale preliminary RL experiments on both the base and reasoning SFT model to guide the design of our training recipe.

\paragraph{Lesson 1: Inefficacy of hyperparameter tuning on proxy models.} To reduce computational overhead, we initially attempt to optimize RL training recipes on an intermediate pre-trained checkpoint (proxy model) before applying them to the full reasoning SFT model. However, we observe that hyperparameters tuned on the proxy model do not reliably generalize to the stronger SFT model, despite the shared architecture. For example, while a specific training recipe yields an approximately 18\% performance gain on AIME 24 when applied to the base model, applying the exact same configuration to the reasoning SFT model resulted in performance stagnation or even degradation. This suggests that the optimal policy update dynamics shift significantly after supervised fine-tuning, necessitating direct tuning on the target model.

\paragraph{Lesson 2: Impact of reward shaping on unparsable trajectories.} Reward formulation is essential to guide policy updates, particularly regarding the handling of format errors. We observe that when the model output cannot be parsed against the ground truth, such instances must be strictly masked out to prevent them from contributing to the gradient. In our experiments, retaining these instances while applying auxiliary reward terms (e.g., length penalty) resulted in non-zero advantages. This introduces significant noise into the training signal, as the policy might be inadvertently incentivized to optimize for auxiliary constraints rather than correct reasoning logic.

\paragraph{Lesson 3: Necessity of difficulty alignment to prevent gradient vanishing.}
We observe that RLFT is highly sensitive to the difficulty distribution of the training data relative to the model's capability. As defined in Equation~\ref{eq:grpo}, the advantage $A_j$ is derived from the relative performance within a group of rollouts. In scenarios where problems are trivially easy or excessively difficult, the intra-group reward variance collapses (i.e., all rollouts receive identical rewards). This results zero advantage ($A_j \rightarrow 0$) and subsequently vanishes the gradients, making the sampling and evaluation costs trivial. This implies that constructing a well-aligned dataset\textemdash where the model succeeds on some rollouts but fails on others\textemdash is essential for effective learning.

\paragraph{Lesson 4: Addressing computational bottlenecks with mixed-policy training.}
RLFT is primarily bottlenecked by the inference cost of generating $n$ rollouts per prompt, whereas the parameter update step negligible in comparison. This computational burden implies that even with large-scale curated datasets, training on the full corpus is often infeasible. Furthermore, we observe that on-policy training exhibits high variance, failing to guarantee monotonic improvement or stable convergence. These limitations emphasize the necessity of leveraging mixed-policy strategies to enhance both sample efficiency and training stability.

\subsection{Recipe}
\paragraph{LLM-as-a-data-filtering.} 

\begin{figure}[!h]
\centering
\definecolor{themeBlue}{RGB}{0, 112, 192}       
\definecolor{themeGreen}{RGB}{84, 130, 53}      
\definecolor{themeOrange}{RGB}{197, 90, 17}     
\definecolor{themeGray}{RGB}{89, 89, 89}        

\resizebox{0.9\textwidth}{!}{%
\begin{tikzpicture}[
    node distance=0.8cm and 1.2cm,
    font=\small\sffamily,
    >=Latex,
    data_node/.style={
        rectangle, draw=themeGreen, fill=white, thick,
        minimum width=2cm, minimum height=0.8cm,
        rounded corners=3pt, align=center,
        font=\sffamily\scriptsize\bfseries
    },
    process_node/.style={
        rectangle, draw=themeBlue, fill=white, thick,
        minimum width=2.0cm, minimum height=1.0cm,
        rounded corners=3pt,
        align=center, font=\sffamily\scriptsize
    },
    decision_node/.style={
        diamond, aspect=2, draw=themeOrange, fill=white, thick,
        minimum width=2.5cm, inner sep=0pt,
        align=center, font=\sffamily\scriptsize
    },
    model_node/.style={
        ellipse, draw=themeGray, fill=white, dashed, thick,
        minimum width=2.0cm, minimum height=0.8cm,
        align=center, font=\sffamily\scriptsize
    },
    line/.style={draw=themeGray, thick, ->, rounded corners=4pt},
    dashline/.style={draw=themeGray, thick, dashed, ->, rounded corners=4pt},
    label_text/.style={font=\scriptsize\color{themeGray}, midway, anchor=center, fill=white, inner sep=2pt}
]

    \node[data_node] (raw_data) {Raw Pool ($\mathcal{D}$)};

    \node[process_node, below=0.8cm of raw_data] (step1) {
        \textbf{Step 1: Sampling}\\
        \scriptsize (Problem Selection)
    };

    \node[process_node, below=0.8cm of step1] (step2) {
        \textbf{Step 2: Rollout}\\
        \scriptsize (Response Generation)
    };
    
    \node[model_node, left=0.8cm of step2] (llm) {LLM ($\mathcal{M}$)};

    \node[process_node, right=1.5cm of step2] (step3) {
        \textbf{Step 3: Scoring}\\
        \scriptsize (Pass Rate Calc.)
    };

    \node[decision_node, above=0.8cm of step3] (step4) {
        \textbf{Step 4: Filter}\\
        \scriptsize (Difficulty Check)
    };

    \node[data_node, above=0.8cm of step4] (final_data) {Aligned Data ($\mathcal{D}'$)};
    
    \node[right=1.0cm of step4, font=\scriptsize\color{gray}, align=left] (discard) {Discard\\(Mismatch)};

    \begin{pgfonlayer}{background}
        \node[fit=(raw_data)(step2)(llm), draw=gray!30, fill=gray!5, rounded corners, dashed, inner sep=0.3cm] (bg_gen) {};
        \node[above right, font=\bfseries\scriptsize\color{themeGray}] at (bg_gen.north west) {Phase I: Generation};

        \node[fit=(step3)(step4)(final_data), draw=gray!30, fill=gray!5, rounded corners, dashed, inner sep=0.3cm] (bg_eval) {};
        \node[above right, font=\bfseries\scriptsize\color{themeGray}] at (bg_eval.north west) {Phase II: Evaluation};
    \end{pgfonlayer}

    \draw[line] (raw_data) -- (step1);
    \draw[line] (step1) -- node[label_text, yshift=3pt] {$x$} (step2);
    \draw[dashline] (llm) -- (step2);

    \draw[line] (step2.east) -- node[label_text, xshift=-2.5pt] {$y_1 \dots y_n$} (step3.west);

    \draw[line] (step3) -- node[label_text, yshift=-3pt] {$\hat{p}(x)$} (step4);
    
    \draw[line] (step4) -- node[label_text, yshift=-3.5pt] {Pass} (final_data);
    \draw[line, color=themeOrange] (step4) -- node[label_text, xshift=-2pt, text=themeOrange] {Fail} (discard);

\end{tikzpicture}
}%
\caption{\textbf{Schematic of the Data Filtering Pipeline.} We denote the initial problem pool as $\mathcal{D}$ and the target LLM as $\mathcal{M}$. In \textbf{Step 1}, a problem $x$ is sampled from $\mathcal{D}$. In \textbf{Step 2}, the model generates $n$ solution rollouts, denoted as $\{y_1, \dots, y_n\} \sim \mathcal{M}(\cdot|x)$. \textbf{Step 3} calculates the pass rate $\hat{p}(x)$ based on correctness. Finally, \textbf{Step 4} applies a difficulty filter, retaining only samples where $\alpha \le \hat{p}(x) \le \beta$ to form the aligned dataset $\mathcal{D}'$.}
\label{fig:filtering_pipeline_refined}
\end{figure}
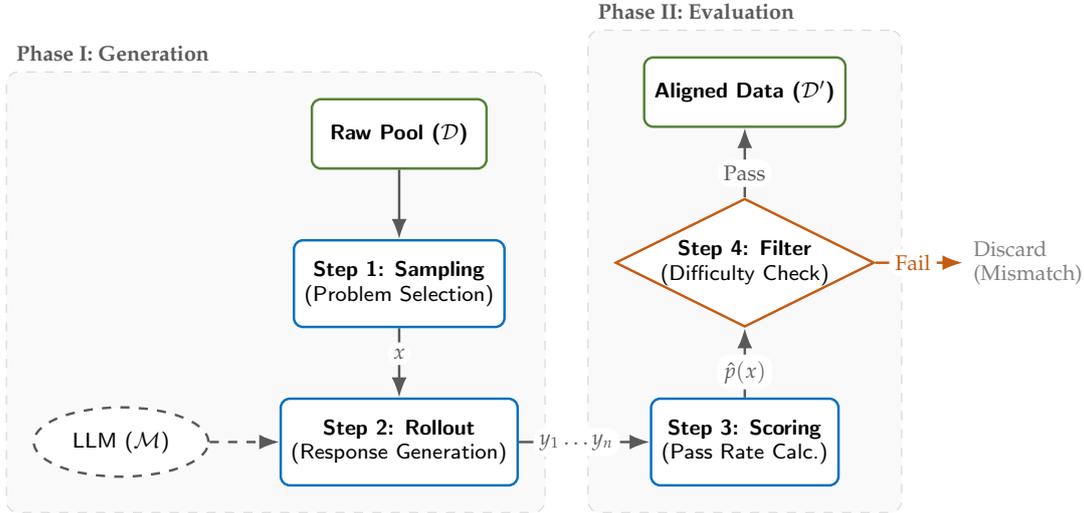

To mitigate zero advantage by difficulty range, DAPO~\cite{yu2025dapoopensourcellmreinforcement} employs the dynamic sampling, however this causes inefficient generation cost while training process. To this end, we build a simple yet dataset filtering pipeline to construct dataset whose difficulty range is well-aligned with the target model's capabilities, namely \textit{LLM-as-a-data-filtering}.

Suppose that we have the initial pool of problems $\mathcal{D}$ and the LLM $\mathcal{M}$ used for filtering. For each problem $x \in \mathcal{D}$, we generate $n$ rollouts and calculate the empirical pass rate as follows:
\begin{align}
y_1, \dots, y_n \sim \mathcal{M}(\cdot \mid x), 
\quad
\hat{p}_k(x) = \frac{1}{n} \sum_{j=1}^{n} \mathbb{1}\big[\textbf{$x$ is solved in the top-$k$ rollouts}\big].
\end{align}
We strictly retain problems whose empirical pass rate falls within a target difficulty range $[\alpha, \beta]$:
\begin{equation}
\alpha \le \hat{p}_k(x) \le \beta.
\end{equation}
This ensures that the difficulty range is well-aligned with the model’s capability, thereby mitigating the instability of early training phase.

\paragraph{Dataset Construction.}
We apply the \textit{LLM-as-a-data-filtering} pipeline to curate a high-quality multi-task dataset composed of mathematical reasoning, code generation, and instruction following. All candidates are evaluated using the checkpoint immediately preceding the RL stage with $n=5$ rollouts.

\begin{itemize}
    \item \textbf{Mathematical Reasoning:} The initial pool is sourced from the \texttt{GURU-92K} dataset~\cite{cheng2025revisitingreinforcementlearningllm}. We first apply a preliminary filter based on the pre-computed Qwen-30B pass rates provided in the dataset, retaining problems with a score of $0 < p \le 10/16$. Subsequently, applying our filtering pipeline, we select instances lying within the difficulty band of $(0, 0.8]$. To address significant domain imbalance with Combinatorics originally accounting for 50\% of the data, we categorize the problems into Number Theory, Combinatorics, Algebra, and Geometry and perform stratified sampling to equalize the distribution across these four sub-domains.

    \item \textbf{Code Generation:} Similar to the mathematical reasoning pipeline, we source code samples from \texttt{GURU-92K} and apply an initial filter based on the provided pass rates ($0 < p \le 10/16$). We then execute our filtering pipeline with identical settings ($n=5$) and randomly subsample the retained instances to align the dataset size with that of the mathematical reasoning corpus, ensuring a balanced task mixture.
    
    \item \textbf{Instruction Following:} We utilize a synthetic dataset of 10,000 samples constructed for instruction following. To prioritize instructions that challenge the model's current capabilities, we applied a stricter upper bound threshold of $\beta=0.4$, retaining only those samples with an empirical pass rate in the range $(0, 0.4]$.
\end{itemize}

\paragraph{Expanding Clipping Range.} To accelerate convergence, we employ a larger cliping range in our training setup:$\epsilon \in [0.28, 0.40]$. Unlike standard settings that enforce a tight trust region, this expanded range allows the policy to deviate more significantly from the reference model when high-advantage signals are present. This strategy strikes a balance between training efficiency and stability, facilitating faster policy improvement without inducing the collapse often seen in unconstrained settings.

\paragraph{Encouraging Long-Context Reasoning.}
Our empirical observations indicate that the capacity for long-context reasoning is fundamental to effective RL fine-tuning. We noted substantial performance improvements when the model was permitted to generate extended reasoning chains. Conversely, applying a length penalty\textemdash a standard practice in general text generation\textemdash proved detrimental, as it inadvertently discourages the model from exploring necessary intermediate reasoning steps. Consequently, we eliminated the length penalty from our reward formulation entirely. To fully accommodate these extended trajectories, we configured our serving infrastructure to maximize the generation length of vLLM, as detailed in \S~\ref{sec:sys_ops}.

\paragraph{Efficiency via Mixed-Policy Trajectory Reuse.}
A primary bottleneck in RLFT is the prohibitive computational cost associated with rollout generation. To mitigate this, we employ a strategy of reusing the same batch of trajectories across multiple gradient optimization steps. Specifically, at outer iteration $k$, we sample a batch of trajectories from the current policy:
\begin{align}
\mathcal{B}_k \sim P(q), \pi_{\theta_k}(o \mid q),
\end{align}
Subsequently, we perform $S$ steps of gradient updates on this fixed batch as follows:
\begin{align}
\theta_{k,s+1}
= \theta_{k,s}
- \eta\,\nabla_\theta \mathcal{L}_{\text{GSPO}}(\theta_{k,s};\,\mathcal{B}_k),
\quad s = 0,\dots,S-1,
\end{align}
where $\theta_{k,0} = \theta_k$ and the updated parameter for the next iteration is defined as $\theta_{k+1} = \theta_{k,S}$.

Theoretically, this procedure initiates as on-policy sampling since $\mathcal{B}_k$ is drawn from $\pi_{\theta_k}$. However, it transitions into an increasingly off-policy regime as $\theta_{k,s}$ drifts away from the behavior policy $\theta_k$ while the trajectories and their computed advantages remain fixed. The number of inner steps $S$ thus serves as a controller for this mixed on-/off-policy dynamic. While standard iterative GRPO~\cite{shao2024deepseekmathpushinglimitsmathematical} shows unstable performance in our preliminary experiments, this trajectory reuse strategy yields substantially more stable optimization while maximizing training efficiency.

\paragraph{Mitigating task regression via multi-task RL.} We observe that RL training focused on a single domain often induces significant performance regression on other tasks. For instance, optimizing the policy exclusively for instruction following tends to degrade capabilities in mathematical reasoning or code generation, even when the base model possesses strong competencies across all areas. To address this, we adopt a multi-task RLFT framework that jointly trains on all target domains\textemdash Math Reasoning, Code Generation, and IFBench\textemdash within a RL loop. Specifically, each mini-batch is constructed as a mixture of these three tasks, utilizing task-specific reward functions while updating a shared policy. This strategy serves as an effective regularizer, mitigating catastrophic forgetting and ensuring robust performance improvements across all downstream benchmarks.

\section{Conclusion}
In this work, we introduced Motif-2-12.7B-Reasoning, a 12.7B parameter model that challenges the assumption that massive scale is required for complex reasoning by achieving parity with frontier models with larger parameters. We demonstrated that a holistic optimization strategy involving system-level efficiency for 64K contexts, distribution-aligned supervised fine-tuning, and a stabilized reinforcement learning recipe is essential for unlocking robust reasoning capabilities under constrained budgets. By releasing our model alongside a transparent and reproducible training methodology, we aim to dismantle the barriers surrounding reasoning-optimized LLMs and empower the open-source community to drive further innovations in accessible high-performance AI.

\bibliographystyle{unsrtnat}
\bibliography{reference}

\clearpage

\section{Appendix}
\label{sec:appendix}

\subsection{Contributions}

All authors are alphabetically sorted by last name.

\textbf{Technical and management leadership}: Sungmin Lee, Junghwan Lim

\textbf{Core contributors}: Dongseok Kim, Taehyun Kim, Eunhwan Park, Jeesoo Lee, Jeongdoo Lee, Junhyeok Lee

\textbf{Contributors}: Wai Ting Cheung, Dahye Choi, Minsu Ha, Jaeheui Her, Jaeyeon Huh, Hanbin Jung, Changjin Kang, Beomgyu Kim, Minjae Kim, Taewhan Kim, Youngrok Kim, Hyukjin Kweon, Haesol Lee, Kungyu Lee, Dongpin Oh, Yeongjae Park, Bokki Ryu, Dongjoo Weon

\end{document}